\documentclass{article}

\usepackage[dblblindworkshop,final]{neurips_2026}
\workshoptitle{Symmetry and Geometry in Neural Representations (NeurReps)}

\usepackage[utf8]{inputenc}
\usepackage[T1]{fontenc}
\usepackage{amsmath,amssymb,amsfonts}
\usepackage{booktabs}
\usepackage{graphicx}
\usepackage{url}
\usepackage{microtype}
\usepackage{xcolor}
\usepackage{placeins}
\usepackage{hyperref}
\hypersetup{hidelinks}

\setcitestyle{numbers,sort&compress}

\title{Native Geometry vs.\ Linear Recoverability: How Alignment Changes Platonic Scaling}

\author{%
  Angus Hoy\\
  EleutherAI SOAR\\
  UniverseTBD\\
  \texttt{angusjwh@gmail.com}\\
  \And
  Austin Lutterbach\\
  EleutherAI SOAR\\
  UniverseTBD\\
  \texttt{austin.lutterbach@proton.me}\\
  \And
  Jasper Chong\\
  University of Queensland\\
  EleutherAI SOAR\\
  \texttt{jasper.chong@uq.edu.au}\\
  \AND
  Aman Kumar\\
  UniverseTBD\\
  EleutherAI\\
  IUCAA\\
  \texttt{kumaramanasci@gmail.com}\\
  \And
  Michael J. Smith\\
  AstroAI\\
  Harvard-Smithsonian Center for Astrophysics\\
  UniverseTBD\\
  EleutherAI\\
  \texttt{mike@mjjsmith.com}\\
}

\begin{document}

\maketitle

\begin{abstract}
Do larger models learn increasingly similar representations of the same world?
The Platonic Representation Hypothesis (PRH) predicts that they should, but
``similar'' can mean native cosine geometry---nearest-neighbour structure
induced by cosine similarity before any learned alignment---or linear
recoverability: correspondence revealed by a supervised linear map.
Using matched Legacy Survey and Hyper Suprime-Cam (HSC) embeddings across five
model families, we find that Ridge alignment strengthens the relationship
between model size and cross-dataset neighbourhood correspondence relative to
native cosine geometry.
The effect appears in all five families and persists when both representations
are reduced to dimension \(256\), so it is not simply that wider embeddings give
alignment more dimensions to work with.
The learned maps remain far from rotations or uniform scaling, and Sparse
Autoencoder (SAE) and Block-Sparse Featuriser (BSF) features give no advantage
over the same alignment applied to dense embeddings.
Model scale thus improves linear recoverability more clearly than native
cosine-neighbourhood correspondence, without establishing convergence toward a
shared native metric geometry.
\end{abstract}

\section{Introduction}\label{sec:intro}

Representations from independently trained models often become more similar with
increasing capability and scale~\citep{huh2024platonic}.
The Platonic Representation Hypothesis (PRH) interprets this as convergence
toward a shared statistical model of reality.
Astronomical surveys provide a useful testbed: the Legacy
Survey~\citep{dey2019legacy} and Hyper Suprime-Cam
(HSC)~\citep{miyazaki2018hsc} observe the same physical objects through
distinct depths, resolutions, and filter sets, and prior work reports increasing
mutual \(k\)-nearest-neighbour (mKNN) agreement between their embeddings as
capacity grows~\citep{duraphe2025platonicuniverse}.

Two representations may become intrinsically more geometrically similar as
models scale, or they may remain geometrically different while becoming easier
to translate between with a learned map.
We distinguish these possibilities.
Prior work studies alignment through supervised linear
maps~\citep{mikolov2013exploiting} and unpaired embedding
translation~\citep{lample2018word,moschella2023relative,jha2025vec2vec}.
Here \emph{native cosine geometry} is the nearest-neighbour structure induced
directly by cosine similarity before any learned alignment, and \emph{linear
recoverability} is the correspondence exposed after fitting a supervised
affine/Ridge map on paired training objects (\emph{linear alignment}).
An affine map can improve neighbourhood correspondence without preserving
distances or angles, so stronger recoverability need not imply a shared native
metric geometry.
Terminology is summarized in Appendix~\ref{app:terminology}.

On the same held-out objects we measure mean \(k\)-nearest-neighbour agreement
before and after linear alignment,
\[
M_{\mathrm{native}}\equiv M_{\mathrm{Dense}},
\qquad
M_{\mathrm{linear}}\equiv M_{\mathrm{Dense+Ridge}},
\]
with alignment gain \(G=M_{\mathrm{linear}}-M_{\mathrm{native}}\)
(Section~\ref{sec:setup}).

We report three main findings.
\textbf{(1)}~Linear recoverability scales more strongly with model size than
native correspondence.
\textbf{(2)}~The effect persists after independently projecting both surveys to
\(256\) principal components, so increased alignment dimensionality alone does
not explain it.
\textbf{(3)}~The maps exposing this additional correspondence remain strongly
anisotropic---scaling different directions by different amounts---while most of
the mKNN gain is carried by a much smaller singular subspace.
Secondary PCA, SAE/BSF~\citep{gao2025sae,fel2026bsf}, same-dataset cross-model,
and unpaired nonlinear controls are reported in the
appendix.\footnote{\url{https://github.com/hypervalis/native-geometry-linear-recoverability}}

\section{Setup}\label{sec:setup}

We use the official UniverseTBD Legacy\(\leftrightarrow\)HSC matched
embeddings~\citep{duraphe2025platonicuniverse} (\(n{=}16{,}384\) objects)
across five families---AstroPTv2~\citep{smith2024astropt},
ConvNeXtv2~\citep{woo2023convnextv2}, DINOv2~\citep{oquab2024dinov2},
ViT~\citep{dosovitskiy2021vit}, and I-JEPA~\citep{assran2023ijepa}---sixteen
size tiers in total (Appendix~\ref{app:details}), with only two tiers for
I-JEPA.

Let \(N_k^A(i)\) and \(N_k^B(i)\) be the \(k\)-nearest-neighbour sets of object
\(i\) under cosine similarity in representations \(A\) and \(B\) (excluding
\(i\)).
Our observable is the mutual \(k\)-nearest-neighbour score
\[
M_k
=
\frac{1}{n}\sum_{i=1}^{n}
\frac{\bigl|N_k^A(i)\cap N_k^B(i)\bigr|}{k},
\]
so \(M_k\) is mean neighbourhood overlap rather than a direct cross-space cosine
comparison~\citep{huh2024platonic,duraphe2025platonicuniverse}; \(M_k{=}1\) for
identical neighbourhoods.
We use \(k{=}10\) as the primary setting and write this as mKNN@10.

All primary results use a frozen train/test split
(\(n_{\mathrm{train}}{=}13{,}107\), \(n_{\mathrm{test}}{=}3277\), seed \(0\))
and a \textbf{test-only gallery}: \(M_k\) is averaged over held-out objects, and
neighbours are sought only among held-out objects, so chance is
\(M_{10}\approx k/(n_{\mathrm{test}}{-}1)\approx0.0031\).

We index model size by \(P\), the approximate parameter count in millions
(Appendix~\ref{app:details}), and write \(X\) and \(Y\) for the paired Legacy
and HSC embedding matrices.
Direction is Legacy\(\to\)HSC\@.
We fit an affine Ridge predictor on paired training objects after train-only
standardization (\(\mathrm{StandardScaler}\) on \(X,Y\), then Ridge with
\(\alpha{=}1\) and an intercept), mapping Legacy into HSC space.
Geometric diagnostics characterise its linear component \(A_{\mathrm{eff}}\) in
original coordinates (Appendix~\ref{app:geometry}).

\section{Linear recoverability scales more strongly than native correspondence}
\label{sec:scaling}

The \emph{level} of cross-survey correspondence and how that correspondence
changes with model size are distinct.
We also test whether sparse feature spaces expose more shared representational
structure by applying the same Ridge alignment to SAE and BSF features.
We compare native dense embeddings (Dense), Ridge-aligned dense embeddings
(Dense+Ridge), and the same Ridge procedure applied to SAE and BSF features
(SAE+Ridge and BSF+Ridge).
Relative to native Dense, SAE+Ridge and BSF+Ridge raise held-out mKNN@10 from
\(0.022\) to \(0.033\) and \(0.035\), but neither consistently steepens the
size slope, and both remain below Dense+Ridge (\(0.042\);
Appendix~\ref{app:sparse}).
The interesting Dense+Ridge result is therefore not merely that Ridge improves
mKNN, but that the improvement grows with model size.

\begin{figure}[t]
\centering
\includegraphics[width=0.88\linewidth]{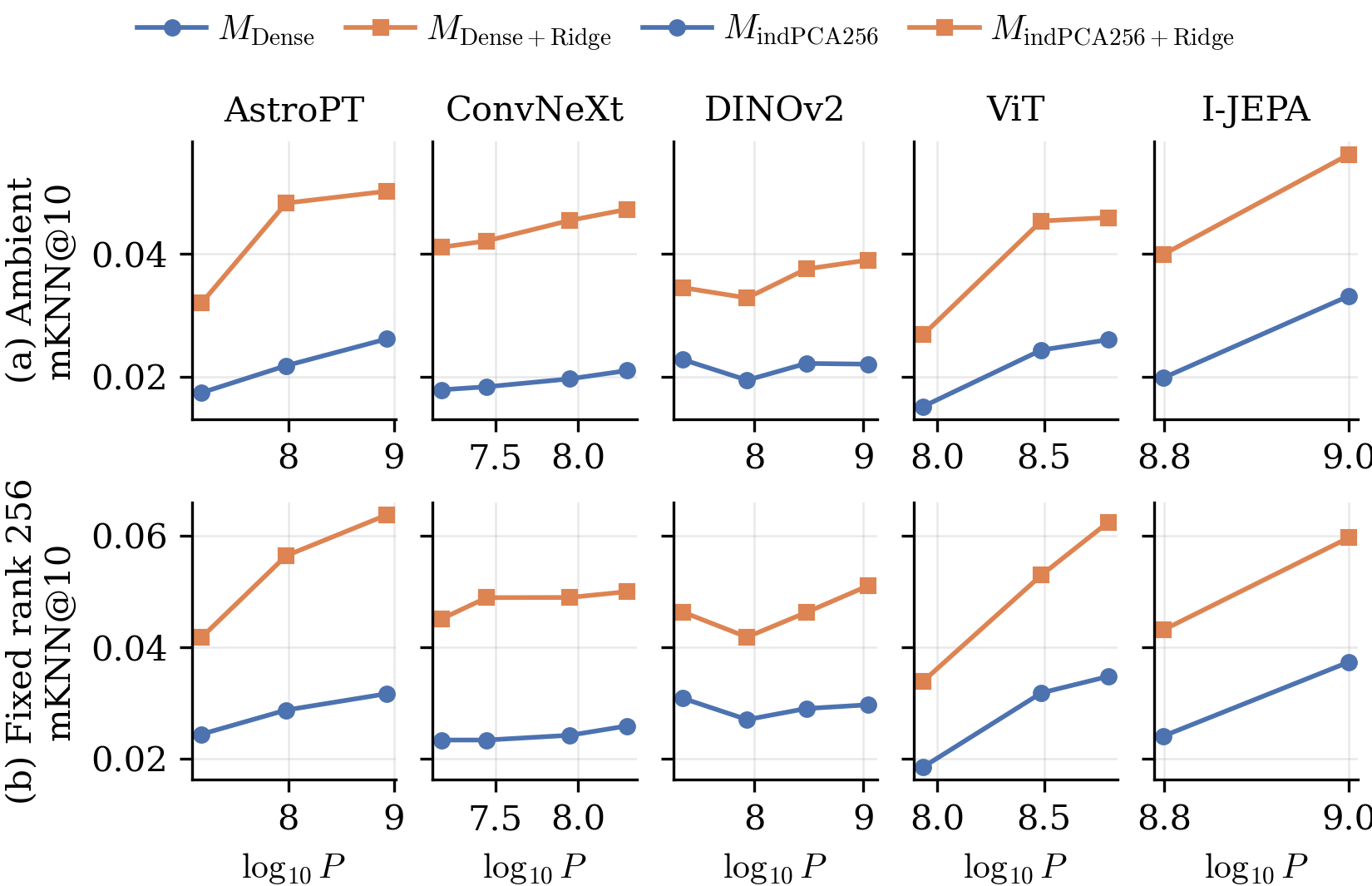}
\caption{
Held-out mKNN@10 versus \(\log_{10}P\) (\(P\): parameter count in millions;
Appendix~\ref{app:details}).
Curves compare native Dense correspondence with Dense+Ridge linear
recoverability.
\textbf{(a)}~Original embedding dimensions: Ridge raises both the level and the
size slope (\(T=0.00721\)).
\textbf{(b)}~Independent PCA to \(256\) dimensions per survey (same feature
dimension in both conditions): the slope amplification persists
(\(T_{256}=0.00864\)).
}
\label{fig:scaling}
\end{figure}

For Legacy\(\to\)HSC, Dense+Ridge raises mean held-out mKNN@10 from \(0.022\)
to \(0.042\) (\(G\approx0.020\); Figure~\ref{fig:scaling}a)---roughly twice
native overlap and about \(13\times\) chance (\(\approx0.0031\)).
For each architecture family \(F\), fitting mKNN against \(\log_{10}P\) gives
\[
\Delta\beta_F
=
\beta_{\mathrm{Dense+Ridge},F}
-
\beta_{\mathrm{Dense},F}.
\]
We find \(\Delta\beta_F>0\) in all five families, and the equal-family mean
\[
T
=
\frac{1}{5}\sum_F\Delta\beta_F
=
0.00721
\]
shows that correspondence exposed by alignment grows faster with size than
native correspondence does.
Averaging families equally makes architecture family, not checkpoint count, the
replication unit.
The effect is positive in all five families (\(p_{\mathrm{sign}}=0.031\)),
remains positive under held-out-object bootstrap (95\% CI
\([0.0034,0.0111]\)), and exceeds a synchronized correspondence-shuffle refit
null (\(p_{\mathrm{perm}}\approx0.005\); Appendix~\ref{app:stats}).
It is stable across \(k\in\{5,10,25,50\}\).
HSC\(\to\)Legacy is weaker and more heterogeneous, so we treat linear
recoverability as directional rather than as a symmetric equivalence
(Appendix~\ref{app:dirk}).

\section{Recoverability is concentrated within broadly anisotropic maps}
\label{sec:geometry}

Scaled-orthogonal maps \(x\mapsto cQx\) preserve pairwise cosine similarities,
so they cannot change cosine mKNN.
The observed gain must therefore come from non-geometry-preserving parts of the
fitted affine map, potentially including translation.

Let \(W\) be the fitted Ridge coefficient matrix and
\(D_x=\mathrm{diag}(\sigma_x)\), \(D_y=\mathrm{diag}(\sigma_y)\) the
train-feature standard deviations.
The effective linear component in original coordinates is
\[
A_{\mathrm{eff}}=D_yWD_x^{-1}=U\Sigma V^\top,
\qquad \Sigma=\mathrm{diag}(s_1,\ldots,s_d),
\]
with singular values \(s_i>0\).
Direction-dependent stretch is summarized by
\[
A_{\log}=\mathrm{std}_{i}(\log s_i),
\]
which ranges from \(1.76\)--\(3.15\) across tiers: alignment requires
substantial anisotropic stretch rather than rotation/reflection and global
rescaling alone.

How broadly is that anisotropy distributed?
With centered log-stretches \(a_i=\log s_i-\tfrac1d\sum_j\log s_j\) and
anisotropy masses \(m_i=a_i^2\), let \(f_{90}^{\mathrm{anis}}\) be the
smallest fraction of singular directions (ordered by decreasing \(m_i\)) needed
for \(90\%\) of \(\sum_i m_i\).
Across the \(16\) maps, \(f_{90}^{\mathrm{anis}}\) is \(0.411\)--\(0.490\)
(median \(0.427\)): departure from uniform scaling is spread across a
substantial fraction of the representation.

\begin{figure}[!t]
\centering
\includegraphics[width=0.72\linewidth,height=0.38\textheight,keepaspectratio]{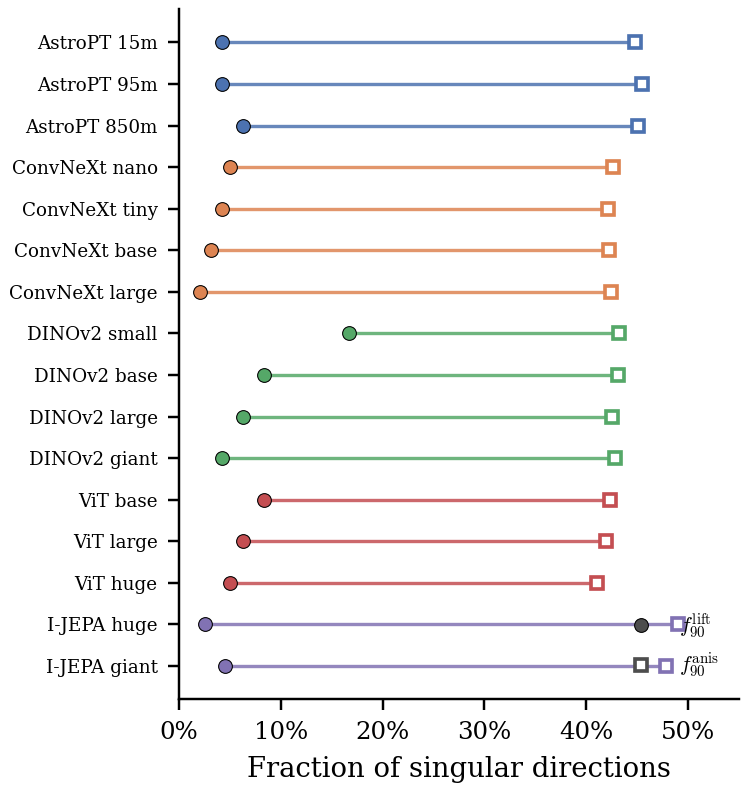}
\caption{
Filled markers: fraction of \(A_{\mathrm{eff}}\) singular directions needed to
recover \(90\%\) of the Dense+Ridge mKNN lift (median \(4.8\%\)).
Open markers: fraction needed for \(90\%\) of the map's anisotropy mass
(median \(42.7\%\)).
The map is anisotropic across many directions, but most neighbour recovery is
concentrated in a much smaller subspace.
}
\label{fig:geometry_concentration}
\end{figure}

Truncating \(A_{\mathrm{eff}}\) along singular directions without refitting
recovers \(90\%\) of the Dense+Ridge mKNN lift
\(G=M_{\mathrm{Dense+Ridge}}-M_{\mathrm{native}}\) with only \(2.1\)--\(17\%\)
of directions (median \(4.8\%\); Figure~\ref{fig:geometry_concentration}).
Spectral anisotropy and functional recoverability therefore have sharply
different effective dimensionalities.

This contrast does not itself show a compelling size trend:
\(f_{90}^{\mathrm{anis}}\) decreases slightly in four of five families, but
only by a few percentage points, and the pooled association with \(\log_{10}P\)
is weak (descriptive Spearman \(\rho{=}0.13\), \(p{=}0.64\), \(n{=}16\));
\(A_{\log}\) and \(f_{90}^{\mathrm{anis}}\) are likewise weakly associated
(\(\rho{=}0.39\), \(p{=}0.13\); Appendix~\ref{app:geometry}).
The part of \(A_{\mathrm{eff}}\) that matters for cosine-neighbour recovery
remains functionally low-dimensional.

\section{Discussion and conclusion}\label{sec:disc}

Apparent representational convergence depends on the transformation class used
to compare representations.
Native cosine correspondence increases only weakly with model size, and
scaled-orthogonal maps leave cosine neighbourhoods unchanged.
Under supervised affine/Ridge alignment the size--correspondence relationship
is stronger (\(T{=}0.00721\); \(5/5\) families) and survives fixed
independent-PCA dimension \(256\) (\(T_{256}{=}0.00864\); \(5/5\)).
The learned maps remain anisotropic, and translation ablation leaves most of the
gain and slope amplification intact, so the effect is carried primarily by
\(A_{\mathrm{eff}}\) rather than by an intercept
(Appendix~\ref{app:geometry}).
Reverse amplification is weaker (\(T_{H\to L}{=}0.00447\)), and a same-dataset
cross-model control retains \(G_{A\to B}\neq G_{B\to A}\)
(Appendix~\ref{app:dirk},\ref{app:crossmodel}).
Limitations include paired Ridge supervision, five families, one astronomy
setting, and a frozen split; \(T\) is stable across Ridge \(\alpha\) and \(k\)
(Appendix~\ref{app:stats},\ref{app:dirk}).
Scaling strengthens supervised linear recoverability more clearly than native
cosine-neighbourhood correspondence, without establishing shared-metric
convergence.

The character of this correspondence is not strictly Euclidean geometric
convergence: two representation spaces may be different perspectives on the same
objects, related by a learned transformation.
While affine Ridge mappings improve mKNN--size scaling substantially, absolute
mKNN remains low.
One possibility is that the scaling relationship becomes substantially stronger
at larger sizes.
Another is that the true recovery map is only weakly approximated by a global
affine map---more complicated or ill-conditioned.
If the underlying relationship is sufficiently smooth, a natural hypothesis is
that within small regions of representation space it can be approximated by a
local affine map, potentially revealing sharper Platonic scaling.
Testing that would help determine whether limited global alignment reflects
genuinely different native geometries or the inadequacy of a single affine
transform.

\section*{Acknowledgements}

We would like to thank Stella Biderman and EleutherAI for their compute and infrastructure support.

{\small
\bibliographystyle{plainnat}
\bibliography{references}
}

\appendix

\section{Terminology and notation}
\label{app:terminology}

\paragraph{Native cosine geometry.}
Nearest-neighbour structure induced directly by cosine similarity in the
original embeddings, before any learned alignment.

\paragraph{Native correspondence.}
Cross-representation agreement measured under native cosine geometry; in the
headline analyses, \(M_{\mathrm{native}}\equiv M_{\mathrm{Dense}}\).

\paragraph{Linear alignment.}
A supervised affine/Ridge map fit on paired training objects that maps one
representation into the other.

\paragraph{Linear recoverability.}
Correspondence exposed after linear alignment; in the headline analyses,
\(M_{\mathrm{linear}}\equiv M_{\mathrm{Dense+Ridge}}\).

\paragraph{Shared metric geometry / isometric convergence.}
The stronger claim that native geometries themselves converge toward the same
metric structure (distances/angles), beyond correspondence under a learned map.

\paragraph{Dense / Dense+Ridge / SAE+Ridge / BSF+Ridge.}
Representation--alignment conditions, not distinct similarity metrics:
\emph{Dense} = unaligned dense embeddings;
\emph{Dense+Ridge} = Ridge on dense embeddings;
\emph{SAE+Ridge} / \emph{BSF+Ridge} = the same Ridge recipe on SAE/BSF features.

\paragraph{mKNN / mKNN@10.}
Mutual \(k\)-nearest-neighbour overlap
\(M_k=\tfrac1n\sum_i|N_k^A(i)\cap N_k^B(i)|/k\);
primary setting \(k{=}10\).

\paragraph{\(A_{\mathrm{eff}}\).}
Effective linear component of the Ridge-aligned map in original coordinates,
\(A_{\mathrm{eff}}=D_yWD_x^{-1}\), where \(W\) is the standardized Ridge
coefficient matrix and \(D_x,D_y\) hold train-feature standard deviations.

\paragraph{Scaled-orthogonal map.}
A map \(x\mapsto cQx\) with \(c>0\) and \(Q^\top Q=I\); preserves pairwise
cosine similarities and therefore cannot change cosine mKNN.

\paragraph{Anisotropy / \(A_{\log}\) / \(f_{90}^{\mathrm{anis}}\).}
Direction-dependent stretching of \(A_{\mathrm{eff}}\):
\(A_{\log}=\mathrm{std}_i(\log s_i)\) for singular values \(s_i\);
\(f_{90}^{\mathrm{anis}}\) is the fraction of singular directions needed for
\(90\%\) of the centered log-stretch mass \(\sum a_i^2\).

\paragraph{\(\Delta\beta_F\) / \(T\).}
For family \(F\), \(\Delta\beta_F=\beta_{\mathrm{Dense+Ridge},F}-\beta_{\mathrm{Dense},F}\)
is the difference of OLS slopes of mKNN versus \(\log_{10}P\);
\(T=\tfrac15\sum_F\Delta\beta_F\) is the equal-family mean.

\section{Protocol and statistical robustness}\label{app:details}\label{app:stats}

Approximate parameter counts (\(10^{6}\)):
AstroPTv2 \(15/95/850\);
ConvNeXtv2 nano/tiny/base/large \(15/28/89/198\);
DINOv2 small/base/large/giant \(22/86/300/1100\);
ViT base/large/huge \(86/307/632\);
I-JEPA huge/giant \(630/1000\).

The primary statistic \(T=\tfrac15\sum_F\Delta\beta_F\) averages family slope differences equally so that architecture family---not checkpoint count---is the replication unit.
The forward effect is positive in all five families (\(p_{\mathrm{sign}}{=}0.03125\)), exceeds a synchronized correspondence-shuffle refit null (\(B{=}200\), \(p{=}1/201\approx0.005\)), and remains positive under held-out-object bootstrap (95\% CI \([0.0034,0.0111]\)); excluding I-JEPA gives \(T_{-I}{=}0.00537\) with four/four families positive.
The result is also stable across \(\alpha\in\{0.01,0.1,1,10,100\}\), with \(T(\alpha)\in[0.00721,0.00848]\) and \(5/5\) positive families throughout.

\section{Direction and neighbourhood-size robustness}\label{app:dirk}

All checks reuse the frozen train/test split, Ridge recipe, and test-only gallery.

\paragraph{Reverse direction.}
HSC\(\to\)Legacy yields \(T_{H\to L}{=}0.00447\) with \(4/5\) families positive (AstroPT negative).
Excluding I-JEPA, \(T_{H\to L}^{(-I)}{=}0.00103\).
Reverse amplification is present but weaker and family-heterogeneous; it is not a second primary claim.

\paragraph{Neighbourhood size.}
Aggregate forward amplification is stable across \(k\in\{5,10,25,50\}\):
\(T(k)\in\{0.00782,0.00721,0.00810,0.00869\}\),
with \(5/5\) positive families at \(k{=}5,10\) and \(4/5\) at \(k{=}25,50\).

\paragraph{Pooled interaction.}
On a tier-by-method table, a family-fixed-effects OLS specification
\(M=\alpha_F+\gamma_F\log_{10}P+\delta_F R+\theta(R\log_{10}P)\), where \(R{=}1\) for Ridge-aligned and \(R{=}0\) for native observations,
gives the same positive interaction direction (\(\theta{=}0.00451\)), providing a descriptive robustness check under a different weighting.
We do not treat ordinary IID OLS uncertainties for this pooled model as primary inference, because Dense/Ridge observations are paired by tier.
Equal-family \(T\), the family sign test, the synchronized shuffle-refit null, and the object bootstrap remain the primary evidence.

\section{Secondary representation controls}\label{app:pca}\label{app:fixedrank}\label{app:sparse}\label{app:residual}

\emph{Independent} PCA+Ridge raises absolute recoverability at mid/high ranks (means \(\approx0.044/0.050/0.055\) at \(128/256/512\)); the fixed-rank \(256\) result is reported in the main text.
A pooled train-only PCA basis without pairing leaves native mKNN near Dense (\(\approx0.024\) at rank \(256\)), showing that dimensional reduction alone does not substantially increase native correspondence.
PCA does not yield a consistent additional scale interaction beyond Dense+Ridge.

Under matched Ridge supervision, SAE and BSF underperform Dense+Ridge by \(0.0085\) and \(0.0062\) mKNN on average, respectively; family-bootstrap 95\% CIs are \([-0.0120,-0.0043]\) and \([-0.0104,-0.0007]\).
Sparse coding therefore provides no additional cross-survey correspondence once alignment supervision is controlled.
SAE protocol: dictionary width \(2048\), seed \(0\), TopK from available checkpoints, train-only Ridge and inverse-document-frequency (IDF) weighting, fixed Legacy\(\to\)HSC~\citep{gao2025sae}; BSF uses Vanilla signed codes with the same Ridge recipe~\citep{fel2026bsf}.
Global Ridge residuals remain weakly locally structured in source space (\(k\in\{10,25,50\}\); all \(16\times 3\) residual-permutation tests give \(p{=}1/201\)), but absolute residual cosine similarity is \(\approx0.01\) and mean relative held-out MSE reduction from a patch intercept is \(\ll0.1\%\); we do not fit local maps.

\section{Geometry details}\label{app:geometry}

\begin{figure}[t]
\centering
\includegraphics[width=\linewidth]{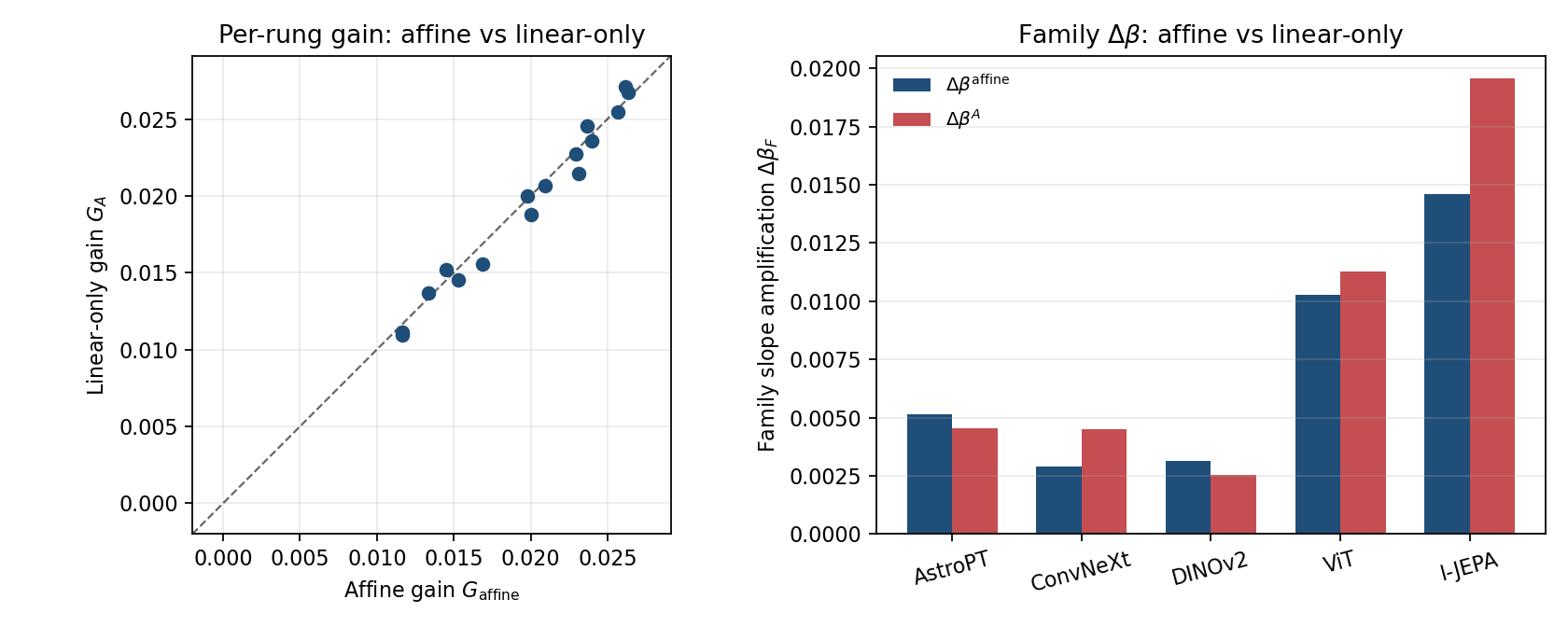}
\caption{
Translation contributes little; the effect is carried by the linear component \(A_{\mathrm{eff}}\).
\textbf{(a)}~Per-tier alignment gain \(G\) over native mKNN, using the full affine map versus the linear component alone with \(b_{\mathrm{eff}}\) removed. Points near the dashed \(y{=}x\) line indicate the translation contributes negligibly (median retained fraction \(\eta_A{=}0.99\)).
\textbf{(b)}~Family slope amplifications \(\Delta\beta_F\) under the two conditions; the model-size amplification survives removal of \(b_{\mathrm{eff}}\) (\(T_A{=}0.00854\) versus \(T_{\mathrm{affine}}{=}0.00721\)).
}
\label{fig:translation_ablation}
\end{figure}

After train-only \(\mathrm{StandardScaler}\), Ridge yields \(W\) and an intercept in standardized coordinates.
We analyze the linear component
\(A_{\mathrm{eff}}=\mathrm{diag}(\sigma_y)W\mathrm{diag}(1/\sigma_x)\)
in original coordinates, where \(\mu_x,\sigma_x\) and \(\mu_y,\sigma_y\) are the train-only per-feature means and standard deviations; the corresponding affine map is \(\hat y=A_{\mathrm{eff}}x+b_{\mathrm{eff}}\) with
\(b_{\mathrm{eff}}=\mu_y-A_{\mathrm{eff}}\mu_x+D_y b_{\mathrm{std}}\), \(D_y=\mathrm{diag}(\sigma_y)\) and \(b_{\mathrm{std}}\) the Ridge intercept in standardized coordinates.
Writing \(A_{\mathrm{eff}}=U\Sigma V^\top\), the distance to the nearest similarity transform is
\(D_{\mathrm{sim}}=\min_{c,Q^\top Q=I}\|A_{\mathrm{eff}}-cQ\|_F/\|A_{\mathrm{eff}}\|_F\)
(main-text values \(D_{\mathrm{sim}}\approx0.88\)--\(0.91\)).
Because cosine mKNN is not translation-invariant, we ablate \(b_{\mathrm{eff}}\) without refitting: for each tier we compare native \(M(X,Y)\), full affine \(M(A_{\mathrm{eff}}X+b_{\mathrm{eff}},Y)\), and linear-only \(M(A_{\mathrm{eff}}X,Y)\) on the same test queries/gallery.
Affine reconstruction matches the sklearn pipeline to \(\max|{\rm err}|\le 3.5{\times}10^{-5}\).
Across the \(16\) tiers the fraction of affine gain retained by the linear component is \(\eta_A=\mathrm{median}\,0.990\) (mean \(0.985\); range \([0.920,1.048]\)).
Family slope amplification without translation is \(T_A{=}0.00854\) (\(5/5\) positive) versus \(T_{\mathrm{affine}}{=}0.00721\), so \(T_A/T_{\mathrm{affine}}{=}1.184\) (excluding I-JEPA, \(T_A^{(-I)}{=}0.00570\)).
A train-mean-centered diagnostic likewise preserves amplification (\(T_{\mathrm{centered}}{=}0.01360\); Figure~\ref{fig:translation_ablation}).
Thus the headline scaling effect is primarily attributable to anisotropic linear deformation rather than affine translation, licensing the geometric analysis of \(A_{\mathrm{eff}}\) below.
On Legacy test \(k_{\mathrm{edge}}\)-NN edges \(\delta_{ij}\), local log-stretch is
\(s_{ij}=\log(\|A_{\mathrm{eff}}\delta_{ij}\|_2/\|\delta_{ij}\|_2)\), summarized by \(\sigma_{\mathrm{local}}=\mathrm{std}(s_{ij})\).
Four of five family slopes of \(\sigma_{\mathrm{local}}\) versus \(\log_{10}P\) are negative, with DINOv2 approximately flat/slightly positive.
Two further diagnostics summarize the singular value spectrum \(\sigma_1\ge\cdots\ge\sigma_d\) of \(A_{\mathrm{eff}}\).
The log-anisotropy \(A_{\log}=\mathrm{std}(\log\sigma_i)\) vanishes exactly when all singular values are equal, i.e.\ when \(A_{\mathrm{eff}}=cQ\) for some \(c>0\) and \(Q^\top Q=I\); it therefore measures departure from \(\mathcal{T}_{\mathrm{scaled\,orth}}\) directly, without the optimal-rotation fit required by \(D_{\mathrm{sim}}\).
The normalized spectral entropy \(H_{\mathrm{norm}}=-\sum_i p_i\log p_i/\log d\), with \(p_i=\sigma_i^2/\sum_j\sigma_j^2\), is \(1\) for a spectrally flat map and decreases as the map concentrates its action on fewer directions.
Both indicate strongly non-flat effective maps at every tier.
Truncating the fitted effective map without refitting shows that roughly \(6\%\) of singular directions recover at least \(90\%\) of the Dense+Ridge mKNN lift.

\section{Same-dataset cross-model control}\label{app:crossmodel}

Cross-survey directionality may reflect survey-information differences, so same-Legacy cross-model alignment tests whether asymmetry persists when the observation channel is fixed.
We evaluate all \(101\) valid unordered cross-family pairs among \(16\) Legacy models (train-only independent PCA256 per model; Ridge \(\alpha{=}1\); mKNN@10; test-only gallery; frozen 80/20 split; both directions fit independently).
For unordered pair \(\{A,B\}\),
\[
G_{\mathrm{sym}}=\tfrac12(G_{A\to B}+G_{B\to A}),
\qquad
A_{\mathrm{dir}}=|G_{A\to B}-G_{B\to A}|,
\]
where \(G_{A\to B}=M^{\mathrm{Ridge}}_{A\to B}-M_{\mathrm{native}}(A,B)\) and likewise for \(G_{B\to A}\).

Mean native mKNN is \(0.050\) (chance \(\approx0.00305\)).
Ridge increases correspondence in both directions for \(101/101\) pairs (mean \(G_{\mathrm{sym}}{=}0.060\); mean \(A_{\mathrm{dir}}{=}0.036\); mean relative asymmetry \(A_{\mathrm{rel}}{\approx}0.33\)).
Mean directional gain is similar for larger\(\to\)smaller (\(\approx0.059\)) and smaller\(\to\)larger (\(\approx0.062\)) maps, so there is no systematic larger-to-smaller advantage.
Descriptively, writing \(\beta_{\min}\) for the OLS slope against \(\log_{10}\) of the \emph{smaller} model's parameter count, both mutual recoverability and directional asymmetry increase with that scale (\(\beta_{\min}(G_{\mathrm{sym}}){=}{+}0.026\), \(\beta_{\min}(A_{\mathrm{dir}}){=}{+}0.017\)); pair-level trends are not treated inferentially because model endpoints recur across pairs.

\section{Fixed-dimensional control.}
A straightforward concern is that larger models often have wider embeddings, and therefore give Ridge a higher-dimensional transformation space in which to fit the correspondence. We test this directly by projecting each survey independently to rank \(r=256\) using train-only PCA. The native and aligned conditions then use exactly the same 256-dimensional features: \(M_{\mathrm{indPCA256}}\) compares them directly, while \(M_{\mathrm{indPCA256+Ridge}}\) applies the same supervised alignment.

Defining
\[
\Delta\beta_{F,256}
=
\beta_{\mathrm{indPCA256+Ridge},F}
-
\beta_{\mathrm{indPCA256},F},
\qquad
T_{256}
=
\frac{1}{5}\sum_F\Delta\beta_{F,256},
\]
we obtain \(T_{256}{=}0.00864\), again with \(\Delta\beta_{F,256}>0\) in all five families
(\(T_{256}^{(-I)}{=}0.00673\) excluding I-JEPA; Figure~\ref{fig:scaling}b).
The scaling effect therefore survives when the alignment dimensionality is held fixed. It is not simply a consequence of larger models giving Ridge more coordinates to work with.

\section{SAE TopK sensitivity}\label{app:sae_topk}

\begin{figure}[t]
\centering
\includegraphics[width=\linewidth]{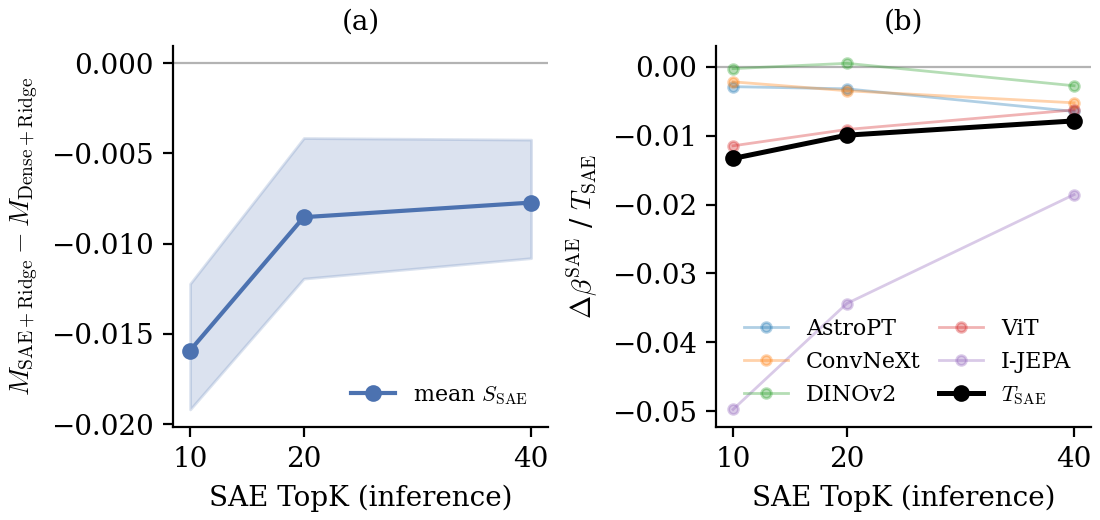}
\caption{
Sparse coding does not overtake the dense baseline at any tested sparsity.
\textbf{(a)}~Mean residual \(M_{\mathrm{SAE+Ridge}}-M_{\mathrm{Dense+Ridge}}\) under inference-time TopK \(K\in\{10,20,40\}\), with family-bootstrap 95\% CIs. Negative values indicate SAE underperforms the matched dense baseline.
\textbf{(b)}~Equal-family scaling interaction \(T_{\mathrm{SAE},K}\), with per-family slope differences \(\Delta\beta^{\mathrm{SAE}}_{F,K}\) faint behind it. Both level and size trend stay non-positive across \(K\).
}
\label{fig:sae_topk}
\end{figure}

Trained checkpoints use heterogeneous TopK in \(18\)--\(23\) with no common training-time \(K\) across all \(16\) tiers.
To test whether the sparse-control result depends on sparsity level, we therefore repeat the frozen Legacy\(\to\)HSC protocol under \emph{inference-time} TopK overrides \(K\in\{10,20,40\}\) on the preferred dictionary width \(2048\) seed-\(0\) checkpoints, holding dictionary width, split, Ridge fitting, mapping direction, IDF, and the test-only gallery fixed.
Across this range the mean residual \(M_{\mathrm{SAE+Ridge}}-M_{\mathrm{Dense+Ridge}}\) stays negative (\(-0.0159\), \(-0.0085\), \(-0.0077\)), family-bootstrap 95\% CIs exclude zero, and the equal-family scaling interaction \(T_{\mathrm{SAE},K}\) remains non-positive (\(-0.0133\), \(-0.0099\), \(-0.0079\); at most one family positive).
Thus the sparse result is not specific to the original TopK choice: across the tested inference sparsity range, SAE does not improve correspondence over the matched Dense+Ridge baseline and does not yield a consistently stronger size trend (Figure~\ref{fig:sae_topk}).

\section{Unpaired nonlinear control}\label{app:unpaired}

An unpaired DualEncoder shared bottleneck (model-specific MLPs into \(Z{=}256\); reconstruction, RBF-MMD, cycle, and Gram terms; no paired cross-survey training objects) yields held-out relational mKNN@10 of \(0.012\)--\(0.023\) versus chance \(\approx0.0035\), but size dependence is non-monotonic on the available ConvNeXtv2 and DINOv2 size tiers.

\end{document}